\documentclass[10pt, a4paper]{article}
\usepackage{lrec}
\usepackage{multibib}
\newcites{languageresource}{Language Resources}
\usepackage{graphicx}
\usepackage{tabularx}
\usepackage{soul}

\usepackage{epstopdf}
\usepackage[utf8]{inputenc}
\usepackage{multicol}
\usepackage{hyperref}
\usepackage{xstring}
\usepackage{comment}

\title{\vspace*{.5\baselineskip} \textbf{{\em gaHealth}: An English--Irish Bilingual Corpus of Health Data}}

\name{Séamus Lankford \textsuperscript{1}, Haithem Afli \textsuperscript{1}, Órla Ní Loinsigh \textsuperscript{2}, Andy Way \textsuperscript{2}} 

\address{\textsuperscript{1} Department of Computer Science, Munster Technological University, Cork, Ireland.   \\
         \textsuperscript{2} Adapt Centre, Dublin City University, Dublin, Ireland \\
         \{seamus.lankford, haithem.afli\}@mtu.ie\\
         \{andy.way, orla.niloinsigh\}@adaptcentre.ie\\}

\abstract{
Machine Translation is a mature technology for many high-resource language pairs. However in the context of low-resource languages, there is a paucity of parallel data datasets available for developing translation models. Furthermore, the development of datasets for low-resource languages often focuses on simply creating the largest possible dataset for generic translation. The benefits and development of smaller in-domain datasets can easily be overlooked. To assess the merits of using in-domain data, a dataset for the specific domain of health was developed for the low-resource English to Irish language pair. Our study outlines the process used in developing the corpus and empirically demonstrates the benefits of using an in-domain dataset for the health domain. In the context of translating health-related data, models developed using the {\em gaHealth} corpus demonstrated a maximum BLEU score improvement of 22.2 points (40\%) when compared with top performing models from the LoResMT2021 Shared Task. Furthermore, we define linguistic guidelines for developing {\em gaHealth}, the first bilingual corpus of health data for the Irish language, which we hope will be of use to other  creators of low-resource data sets. {\em gaHealth} is now freely available online and is ready to be explored for further research. \\ \newline }
\begin{document}
\maketitleabstract

\section{Introduction}
\label{sec:intro}

Improvements in performance in natural language processing (NLP) tasks are typically to be seen when deep learning models are used. However, deep learning requires large amounts of data for model training. Consequently, the availability of large amounts of textual data has become fundamental to the success of NLP applications, such as language modelling \cite{buck2014n} and Neural Machine Translation (NMT) \cite{bahdanau2014neural,sennrich2016edinburgh}.

A popular method of developing such corpora for Machine Translation (MT) tasks is to crawl and parse bilingual web pages to general parallel corpora. However, given the nature of low-resource languages, there are often insufficient web sites available in both the languages of study. Accordingly, a lack of web content typically hinders the development of NLP applications for low-resource languages.
 
The motivation underpinning our present work comes from the challenges we faced in developing high-performing MT models in low-resource settings \cite{afli2017sentiment}. In this work, we developed the first bilingual corpus of  health data for the English-Irish pair. A procedure was created to extract, clean and select appropriate sentences to build a bilingual corpus. In addition, we built a high-performing MT model for translating in-domain health data.

\section{Related work}
\label{sec:relWork}

\subsection{Transformer}

Transformer \cite{vaswani2017attention} is an architecture for transforming an input sequence into an output  sequence  via an Encoder and Decoder without relying on Recurrent Neural Networks. Transformer models use attention to focus on previously generated tokens. This approach allows models to develop a long memory which is particularly useful in the  domain of language translation. 

\subsection{Transformer Hyperparameter Optimization}

Hyperparameter optimization of Transformer models in translating the low-resource English-Irish language pair has been evaluated in previous studies \cite{lankford2021transformers}. 
Carefully selecting the appropriate subword model has been shown to be an important driver of translation performance. A Transformer architecture, using a 16k BPE SentencePiece subword model, demonstrated optimal performance. 

\subsection{Neural Machine Translation}
Using large bilingual corpora, NMT approaches require the training of neural networks to learn a statistical model for machine translation. The technique has demonstrated state-of-the-art translation performance on many benchmarks. However, one of the key factors in enabling the development of high performing NMT models is the availability of large amounts of parallel data \cite{koehn2017six,sennrich2019revisiting}.

\section{Proposed Approach}
\label{sec:meth}
\subsection{Sources for {\em gaHealth} Development}
\label{subsec:signCorp}
To build a bilingual corpus of health data, we  selected multiple sources of professionally translated documents from within the Irish government, all of which are publicly available. In particular, the bilingual strategy statements and annual reports of the Irish Department of Health since 2010 were chosen. 

Furthermore, a dataset of Covid-related data, developed for a previous study~\cite{lankford2021machine}, was incorporated into a larger health dataset. Given the pace at which Covid-19 data was being published, translated Irish website content often lagged the English-language counterpart. Website snapshots taken by the WayBack Machine \footnote{\url{https://archive.org/web/}} \cite{arora2016using} proved particularly useful in creating good parallel data from unaligned parallel websites.

This amalgamated corpus, {\em gaHealth}, consists of 16,201 lines of parallel text files. The combined English and Irish vocabulary size is 19,269 unique words. The constituent elements of the dataset, prior to applying the toolchain, are outlined in Table \ref{tab:elements}.

\begin{table}
\centering
\begin{tabular}{llllll}
\hline
\textbf{Documents} &
 \textbf{Source} &
 \textbf{Lines} & \\ \hline
Strategy Statement 2020 & HSE & 3k &  \\
Strategy Statement 2017 & HSE & 2.5k & \\
Strategy Statement 2015 & HSE & 3k &  \\
Annual Report 2020 & HSE & 2k &  \\
Annual Report 2019 & HSE & 2k &  \\
Annual Report 2017 & HSE & 2k &  \\
Website (Covid) & Citizen's Advice & 4k & \\
Publications (Covid) & HSE & 4k \\ \hline
\end{tabular}

\caption{Sources used in corpus development}
\label{tab:elements}
\end{table}

\subsection{Toolchain used for {\em gaHealth} Development}
\par

\label{subsec:creatCorp}

The HSE PDF and Word documents were pre-processed using a toolchain currently under development as part of the Irish Language Resource Infrastructure project (ILRI), funded by the Department of the Gaeltacht. This toolchain has been written to accept primarily data that originates in public administration organizations, i.e. relatively formal text for which the translation quality is assumed to be high, the structure/formatting to be reasonably consistent, and the potential for noise to be low.

The source material consisted of a combination of twelve input files: six in English (all PDF) and six in Irish (five PDF, one Word); all PDFs had a text layer. They ranged from relatively short (30-40 A4 pages) to more substantial (ca. 200 pages). PDFs in particular can be problematic for creating high-quality corpora for a variety of reasons. In other words, while the quality of the input content in this case can be said to be high, the quality of the input medium is low.

\begin{figure*}
\includegraphics [width=1\textwidth]{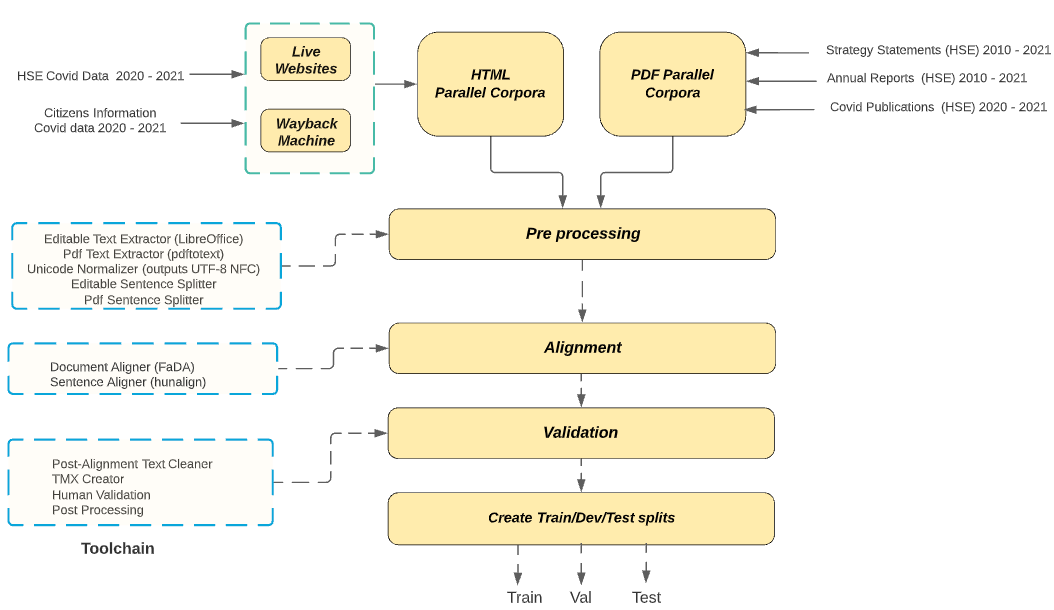}
\caption{Corpus development process.  In developing the corpus, the key steps of data collection, pre-processing, alignment and validation were followed. The role of the toolchain at various stages is highlighted.}
\label{fig:corpus_dev}
\end{figure*}

The process used for developing the corpus is illustrated in Figure \ref{fig:corpus_dev}. The toolchain consists of a set of components run in sequence over a set of input documents, in order to convert it from raw content to a sentence-aligned corpus. Several of the components listed below have different implementations depending on the source type and intended output.

\paragraph{Text extractors}These are wrappers around external components that extract text based on input type; for the {\em gaHealth} file types, wrappers were written for LibreOffice and pdftotext.

\paragraph{Unicode normalizer}This is used to achieve Unicode equivalence, and optionally to substitute certain (e.g. corrupted) characters.

\paragraph{Language detector}In this toolchain, the language detector is a wrapper around langdetect, itself a port from language-detection by Nakatani Shuyo~\cite{nakatani2010langdetect}. The wrapper was written to allow this to be run conveniently either on a string or on an entire file.

\paragraph{Sentence splitters}These are custom-written components to reconstruct sentence boundary information. For editable file types like plain text and Word, this process is relatively straightforward. However, PDFs present particular challenges in this regard. Along with ordering issues, the absence of sentence boundary information is one of the biggest reasons why it is so difficult to construct a high-quality corpus from PDFs. A custom sentence splitter was used to determine sentence boundaries from text extracted from PDFs specifically, primarily using capitalization and language-specific lists of abbreviations to determine where sentences should be broken.

\paragraph{Document aligner}This aligns sets of files whose languages have been identified. A wrapper was written around an external component called FaDA~\cite{lohar2016fada} to adapt it to the toolchain. As FaDA always names an alignment for each input file, sometimes even mapping two different files to the same one, it was necessary to put some selection logic in here, as well as a mechanism for determining when there is no appropriate mapping. Constraints may be put on the relative size of the files to accept an alignment, and the process may be re-run multiple times, with previously rejected files being run again.

\paragraph{Sentence aligner}This aligns pairs of files at the sentence level. This was a wrapper around the external component hunalign~\cite{varga2005hunalign}. No special parameters were used, as default settings produced results of high quality.

\paragraph{Text cleaners}These remove sentence pairs that are believed to be incorrect alignments, such as empty segments and those with obviously mismatched content.

\subsection{Guidelines}
\label{subsubsec:guide}

With the above considerations in mind, the following set of rules was decided upon when processing the {\em gaHealth} dataset. Many of these could be specified as parameters to the toolchain, while others were hard-coded into the system.

\begin{enumerate}

\item Unicode standard: normalize all characters to Unicode UTF-8 NFC. Remove any byte order marks.

\item Whitespacing and capitalization: merge sequences of whitespace characters into a single space. Do not perform tokenization or truecasing.

\item File language detection: scan the first 50 lines, and then every 100th line.

\item Document alignment: assume that specific patterns like a line beginning with a single letter in parentheses or a number followed by a full stop indicate a sentence break from the previous line. Ensure each document is 0.75-1.33 times the size of the document it is being aligned with. Run for a maximum of three iterations.

\item Sentence alignment: allow one-to-many alignments.

\item Cleaning: remove any pairs where source or target:
\begin{itemize}
\item is empty
\item contains no non-alphabetical characters
\item is of an incorrect language. This will remove most untranslated segments. The language is only to be detected for segments that have at least 40 characters
\end{itemize}

\end{enumerate}

\subsection{Transformer architecture}

All EN-GA and GA-EN models, trained with {\em gaHealth}, were developed using a Transformer architecture. Optimal hyperparameters were selected in line with the findings of our previous studies \cite{lankford2021transformers}. These hyperparameters are outlined in in Table \ref{tab:hpo-table}.
\begin{center}
\begin{table}
\center
\begin{tabular}{ll}
\hline
\textbf{Hyperparameter} & \textbf{Values}                \\ \hline
Learning rate            & 0.1, 0.01, 0.001, \textbf{2}            \\ \hline
Batch size               & 1024, \textbf{2048},  4096, 8192       \\ \hline
Attention heads          & \textbf{2}, 4, \textbf{8}                     \\ \hline
Number of layers         & 5, \textbf{6}                           \\ \hline
Feed-forward dimension   & \textbf{2048}                           \\ \hline
Embedding dimension      & 128, \textbf{256}, 512                  \\ \hline
Label smoothing          & \textbf{0.1}, 0.3                       \\ \hline
Dropout                  & 0.1, \textbf{0.3}                       \\ \hline
Attention dropout        & \textbf{0.1}                            \\ \hline
Average Decay            & 0, \textbf{0.0001}                      \\ \hline
\end{tabular}
\caption{Hyperparameter optimization for Transformer models. Optimal parameters are highlighted in bold.}
\label{tab:hpo-table}
\end{table}
\end{center}

\section{Empirical Evaluation}
\label{sec:exp}

In addition to developing the {\em gaHealth} corpus, the effectiveness of the dataset was  evaluated by training models for English-Irish and Irish-English translation in the Health domain. All experiments involved concatenating source and target corpora to create a shared vocabulary and a shared SentencePiece~\cite{kudo2018SentencePiece} subword model.  The impact of using separate source and target subword models was not explored.

To benchmark the performance of our models, the EN-GA and GA-EN test datasets from the LoResMT2021 Shared Task ~\cite{ojha2021findings} were used. These test datasets enabled the evaluation of the {\em gaHealth} corpus, and associated models, since this shared task focused on an application of the health domain i.e. the translation of Covid-related data. Furthermore, using a shared task test dataset enables the comparison of gaHealth models' performance with models entered by other teams. 

The results from the IIITT~\cite{puranik2021attentive} and UCF~\cite{chen2021ucf} teams are included in Tables \ref{tab:en2ga} and \ref{tab:ga2en} so  the performance of the {\em gaHealth} models can be easily compared with the findings of LoResMT2021. IIITT fine-tuned an Opus MT model from Helsinki NLP on the training dataset.  UCF~\cite{chen2021ucf} used transfer learning, unigram and subword segmentation methods for English–Irish and Irish–English translation.

\subsection{Infrastructure}
Rapid prototype development was enabled through a Google Colab Pro subscription using NVIDIA Tesla P100 PCIe 16 GB graphic cards and up to 27GB of memory when available~\cite{bisong2019google}. Our MT models were trained using the Pytorch implementation of OpenNMT 2.0, an open-source toolkit for NMT~\cite{klein2017opennmt}. 

\subsection{Metrics}

Automated metrics were used to determine the translation quality. All models were trained and evaluated using the BLEU~\cite{papineni2002BLEU}, TER~\cite{snover2006study} and ChrF~\cite{popovic2015ChrF} evaluation metrics. Case-insensitive BLEU scores, at the corpus level, are reported. Model training was stopped after 40k training steps or once an early stopping criteria of no improvement in validation accuracy for four consecutive iterations was recorded.

\subsection{Results: Automatic Evaluation}

The hyperparameters used for developing the models are outlined in Table \ref{tab:hpo-table}. The details of the train, validation and test sets used by our NMT models are outlined in Table \ref{tab:en2ga-stats} and Table \ref{tab:ga2en-stats}. In all cases, 502 lines were used from the LoResMT2021 validation dataset whereas the test dataset used 502 lines for EN-GA translation and 250 lines for GA-EN translation. Both were independent health-specific Covid test sets which were provided by LoResMT2021. There was one exception, due to a data overlap between test and training data, a reduced test set was used when testing the {\em gaHealth} en2ga* system.  

\begin{table} [ht]
\centering
\begin{tabular}{lcccccc}
\hline
\textbf{Team} &
 \textbf{System} &
  \textbf{Train}  &
  \textbf{Dev}  &
  \textbf{Test}  \\ \hline
adapt & covid\_extended & 13k & 502 & 500 \\
adapt & combined\_domains & 65k & 502 & 500 \\
IIITT  & en2ga-b & 8k & 502 & 500 \\
UCF     & en2ga-a & 8k & 502 & 500   \\ 
{\em gaHealth} & en2ga & 24k & 502 & 500 \\
{\em gaHealth} & en2ga* & 24k & 502 & 338 \\
\hline
\end{tabular}
\caption{EN-GA Train, Dev and Test dataset distributions. The baseline {\em gaHealth} system was augmented with an 8k Covid dataset provided by LoResMT2021. A smaller test set was used when evaluating {\em gaHealth} en2ga* due to an overlap with the training data. An alternative approach of removing the overlap from the {\em gaHealth} corpus, prior to training, was also carried out to produce the {\em gaHealth} en2ga system.} 
\label{tab:en2ga-stats}
\end{table}

\begin{table} [ht]
\centering
\begin{tabular}{lcccccc}
\hline
\textbf{Team} &
 \textbf{System} &
  \textbf{Train}  &
  \textbf{Dev}  &
  \textbf{Test}  \\ \hline
IIITT  & ga2en-b & 8k & 502 & 250 \\
UCF     & ga2en-b & 8k & 502 & 250   \\ 
{\em gaHealth} & ga2en & 24k & 502 & 250 \\
\hline
\end{tabular} 
\caption{GA-EN Train, Dev and Test dataset distributions. The baseline {\em gaHealth} system was augmented with an 8k Covid dataset provided by LoResMT2021. All overlaps were removed from the {\em gaHealth} corpus prior to training the {\em gaHealth} ga2en model.}
\label{tab:ga2en-stats}
\end{table}

\begin{figure}[h]
    \centering
    {\includegraphics[width=3.95cm]{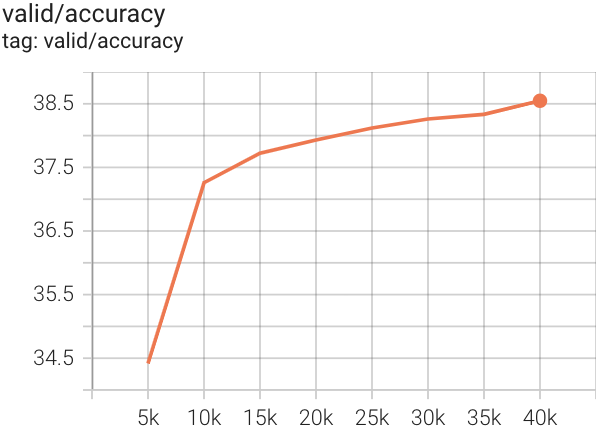}}
    {\includegraphics[width=3.95cm]{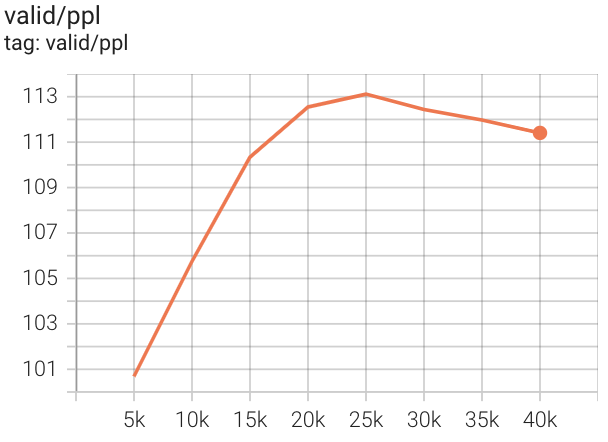}}
    \caption{{\em gaHealth} en2ga* system: training \textit{EN-GA} model with combined 16k gaHealth corpus and 8k LoResMT2021 covid corpus achieving a max validation accuracy of 38.5\% and perplexity of 111 after 40k steps. BLEU score: \textbf{37.6}. }
    \label{fig:en-ga-gaHealth}
\end{figure}

\begin{figure}[h]
    \centering
    {\includegraphics[width=3.95cm]{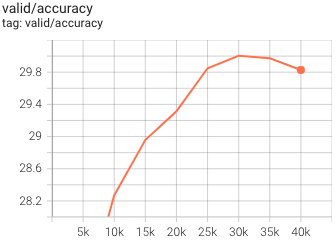}}
    {\includegraphics[width=3.95cm]{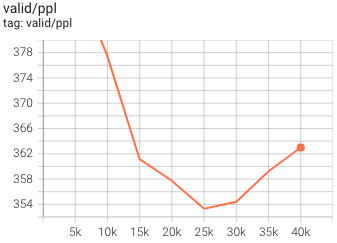}}
    \caption{ adapt covid\_extended system: training \textit{EN-GA} model with 8k LoResMT2021 covid corpus achieving a max validation accuracy of 30.0\% and perplexity of 354 after 30k steps. BLEU score: \textbf{36.0}.}
    \label{fig:en-ga-covid}
\end{figure}

\begin{figure}[h]
    \centering
    {\includegraphics[width=3.95cm]{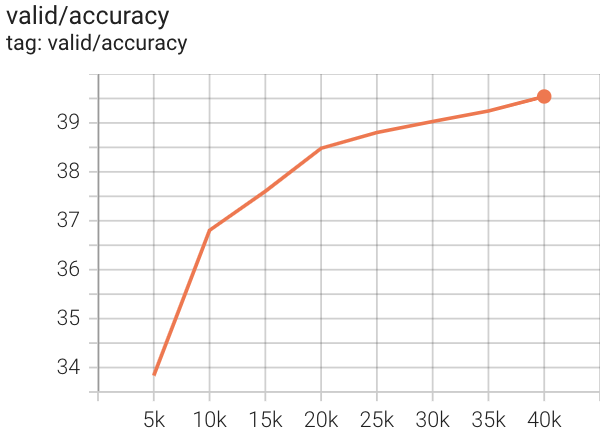}}
    {\includegraphics[width=3.95cm]{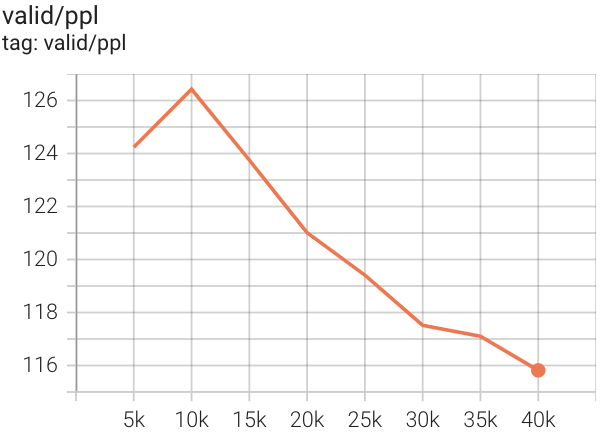}}
    \caption{{\em gaHealth} ga2en system: training \textit{GA-EN} model with combined 16k gaHealth corpus and 8k LoResMT2021 covid corpus achieving a max validation accuracy of 39.5\% and perplexity of 116 after 40k steps. BLEU score: \textbf{57.6.}}
    \label{fig:ga-en-gaHealth}
\end{figure}

\par 
Experimental results achieved using a Transformer architecture, are summarized in Table \ref{tab:en2ga} and Table \ref{tab:ga2en}. In the LoResMT2021 Shared Task, the highest-performing EN-GA system was submitted by the ADAPT team. The system uses an extended Covid dataset (13k, which is a combination of the MT summit Covid baseline and a custom DCU Covid dataset. This Transformer model, with 2 heads, performs well across all key translation metrics (BLEU: 36.0, TER: 0.531 and ChrF3: 0.6). 
\begin{table}[ht!]
\centering
\begin{tabular}{lcccccc}
\hline
\textbf{Team} &
 \textbf{System} &
  \textbf{BLEU} $\uparrow$ &
  \textbf{TER} $\downarrow$ &
  \textbf{ChrF3} $\uparrow$ \\ \hline
UCF     & en2ga-b & 13.5 & 0.756 & 0.37   \\
IIITT  & en2ga-b & 25.8 & 0.629 & 0.53 \\
adapt & combined & 32.8 & 0.590 & 0.57 \\
{\em gaHealth} & en2ga & 33.3 & 0.604 & 0.56 \\
adapt & covid\_extended & 36.0 & 0.531 & 0.60 \\
{\em gaHealth} & en2ga* & \textbf{37.6} & 0.577 & 0.57 \\  
\hline
\end{tabular}
\caption{EN-GA {\em gaHealth} system compared with LoResMT 2021 EN-GA systems.}
\label{tab:en2ga}
\end{table}

\par
Validation accuracy, and model perplexity, in developing the {\em gaHealth} models are illustrated in Figure \ref{fig:en-ga-gaHealth} and Figure \ref{fig:ga-en-gaHealth} whereas Figure \ref{fig:en-ga-covid} illustrates model training on just the covid\_extended dataset. Rapid convergence was observed while training the {\em gaHealth} models such that little accuracy improvement occurs after 30k steps. Only marginal gains were achieved after this point and it actually declined in the case of the system trained using the covid\_extended dataset. 

Perplexity (PPL) shows how many different, equally probable words can be produced during translation. As a metric for translation performance, it is important to keep low scores so the number of alternative translations is reduced. 

\par
Of the models developed by the ADAPT team, the worst-performing model uses a larger 65k dataset. This is not surprising given the dataset is from a generic domain of which only 20\% is health related. The performance of this higher-resourced 65k line model lags the augmented {\em gaHealth} model which was developed using just 24k lines. 

\begin{table}[ht!]
\centering
\begin{tabular}{lcccccc}
\hline
\textbf{Team} &
 \textbf{System} &
  \textbf{BLEU} $\uparrow$ &
  \textbf{TER} $\downarrow$ &
  \textbf{ChrF3} $\uparrow$ \\ \hline
UCF & ga2en-b & 21.3 & 0.711 & 0.45\\
IIITT  & ga2en-b & 34.6 & 0.586 & 0.61\\
{\em gaHealth} & ga2en & \textbf{57.6} & 0.385 & 0.71\\
\hline
\end{tabular} 
\caption{GA-EN {\em gaHealth} system compared with LoResMT 2021 GA-EN systems.}
\label{tab:ga2en}
\end{table}

\par
For translation in the GA-EN direction, the best performing model for the LoResMT2021 Shared Task was developed by IIITT with a BLEU of 34.6, a TER of 0.586 and ChrF3: 0.6. This effectively serves as the baseline by which our GA-EN model, developed using the {\em gaHealth} corpus, can be benchmarked. The performance of the {\em gaHealth} model offers an improvement across all metrics with a BLEU score of 57.6, a TER of 0.385 and a CHrF3 result of 0.71. In particular, the 40\% improvement in BLEU score is very significant.

\section{Discussion}
\label{sec:res}
Although the main objective of this work is to develop the first bilingual corpus of English--Irish data, we conduct initial experiments on the effectiveness of such datasets in training MT models. We have used our {\em gaHealth} dataset to train an MT model on test data from the LoResMT2021 Shared task to evaluate how the system performs translating health data in both the EN-GA and GA-EN direction. Our systems, developed using the {\em gaHealth} corpus achieved significantly higher scores. 

\section{Conclusion and Future Work}
\label{sec:concFut}
The main contribution of this work is to present an ongoing translation project that aims at building the first ever parallel corpus of health data for the Irish language -- {\em gaHealth} -- by fully utilising freely available parallel documents. 

Due to the issues encountered during conversion of PDF documents, we developed guidelines in order to aid in the conversion process. In addition to developing the {\em gaHealth} corpus, we trained and evaluated translation models for in-domain health data.  

In our experiments, the models achieved a BLEU score of 37.6 (Table \ref{tab:en2ga}) for translating EN-GA test data and 57.6 (Table \ref{tab:ga2en}) for translating in the GA-EN direction, which is encouraging performance given this is the beginning of our work on {\em gaHealth}. There is no such corpus available according to the best of our knowledge, so {\em gaHealth} will become a useful resource in the NLP community, especially for those working with the Irish language domain. 

For future work, we intend to extend the corpus, as more Irish language documents become available. Upon extension we will refine our models. One important aspect which needs further investigation is to understand why the EN-GA model ({\em gaHealth} en2ga), tested with the full test set, performed worse than the model ({\em gaHealth} en2ga*) which was tested with the reduced test set. A deep linguistic investigation involving a sentence level BLEU analysis will be conducted as part of a future study. 

In addition, we aim to build in-domain datasets for other key domains such as Education and Finance. We will also apply deep learning techniques to further refine our in-domain models. We have released the {\em gaHealth} corpus online\footnote{\url{https://github.com/seamusl/gaHealth}} to facilitate further research on this data set.

\section{Acknowledgements}
\label{sec:ack}
This research is supported by Science Foundation Ireland through ADAPT Centre (Grant 13/RC/2106)
(www.adaptcentre.ie) at Dublin City University. This research was also funded by the Munster Technological University and the National Relay Station (NRS) of Ireland.

\section{Bibliographical References}
\label{sec:ref}
\bibliographystyle{lrec}
\bibliography{lrec2020W-xample-kc}

\end{document}